\documentclass[letterpaper, 10 pt, conference]{ieeeconf}
\IEEEoverridecommandlockouts
\pdfoutput=1
\usepackage{graphicx} 
\usepackage{caption,subcaption}
\usepackage{scalerel}
\usepackage[table]{xcolor}

\usepackage{times}

\usepackage{graphicx}
\graphicspath{{figures/}}

\usepackage{amsmath}
\usepackage{amssymb}
\usepackage{tabularx}
\usepackage[utf8]{inputenc}
\usepackage[T1]{fontenc}
\usepackage{textcomp}
\usepackage{gensymb}
\usepackage{multirow}
\usepackage{etoolbox}
\usepackage{adjustbox}

\usepackage{soul}

\usepackage{float}
\usepackage{algorithm}
\usepackage[noend]{algpseudocode}
\newfloat{algorithm}{t}{lop}

\usepackage{multicol}

\usepackage[english]{babel}
\usepackage{blindtext}

\usepackage{amsmath}
\usepackage{amssymb}
\usepackage{booktabs}
\usepackage[noadjust]{cite}

\usepackage{float}
\usepackage{fontawesome}
\usepackage{graphicx}
\usepackage{marvosym}
\usepackage{multirow}
\usepackage{xspace}
\usepackage{wrapfig}
\usepackage{xcolor}
\usepackage{soul}

\newcommand{\cf}{\emph{cf.}}

\newcommand{\ie}{\emph{i.e.}}
\newcommand{\eg}{\emph{e.g.}}
\newcommand{\parahead}[1]{\noindent\textbf{#1}:\ }

\newcommand{\filluptopage}[1]{%
  \clearpage
  \loop\ifnum\value{page}<#1\relax
    \null\clearpage
  \repeat
  \loop\ifnum\value{page}=#1\relax
    \null\clearpage
  \repeat
}

\makeatletter
\def\blfootnote{\xdef\@thefnmark{}\@footnotetext}
\makeatother

\usepackage[bookmarks=true, colorlinks=true, pagebackref]{hyperref}

\begin{document}

\newcommand{\coolname}{DRACO\xspace}
\title{\textbf{\coolname: Weakly Supervised Dense Reconstruction And Canonicalization of Objects}}

\author{
Rahul Sajnani$^{*1}$, AadilMehdi Sanchawala$^{*1}$, Krishna Murthy Jatavallabhula$^{2}$, \\Srinath Sridhar$^{3}$, and K. Madhava Krishna$^{1}$
%
\thanks{$^{1}$Robotics Research Center, KCIS, International Institute of Information Technology, Hyderabad, India.
}
\thanks{$^{2}$Mila, Universite de Montreal, Canada.}
\thanks{$^{3}$Brown University, USA.}%
\thanks{$^{*}$Equal contribution by the authors.}
\thanks{\textbf{Please visit our project page \href{https://aadilmehdis.github.io/DRACO-Project-Page/}{here}.}}
}
%

\maketitle

\begin{abstract}
We present \coolname, a method for \underline{D}ense \underline{R}econstruction \underline{A}nd \underline{C}anonicalization of \underline{O}bject shape from one or more RGB images.
Canonical shape reconstruction---estimating 3D object shape in a coordinate space canonicalized for scale, rotation, and translation parameters---is an emerging paradigm that holds promise for a multitude of robotic applications.
Prior approaches either rely on painstakingly gathered dense 3D supervision, or produce only sparse canonical representations, limiting real-world applicability.
\coolname{} performs dense canonicalization using only \emph{weak supervision} in the form of camera poses and semantic keypoints at train time.
During inference, \coolname{} predicts dense object-centric depth maps in a canonical coordinate-space, solely using one or more RGB images of an object.
Extensive experiments on canonical shape reconstruction and pose estimation show that \coolname{} is competitive or superior to fully-supervised methods.
\end{abstract}


\section{Introduction}
%

To execute meaningful real-world tasks, robots must perceive and understand the objects that surround them. Manipulation~\cite{manipulationbook,kemp2007challenges,edsinger2007robot}, autonomous driving~\cite{campbell2010autonomous, thrun2006stanley, dsdnet}, and human-robot interaction~\cite{goodrich2008human} are just a few examples where robots must reason about previously unseen instances of objects~\cite{manuelli2019kpam,wang20206_6pack,burchfiel2017bayesian}.

Recently, \emph{canonicalization} --- the process of mapping an object instances to a category-level container --- has emerged as a useful tool for category-level understanding~~\cite{nocs,c3dpo,articulatedpose,xnocs} (Fig.~\ref{fig:nocs}).
Canonicalization reduces intra-category variation, enabling techniques trained on a handful instances to generalize to the vast majority of previously unseen instances~\cite{kar2015cvpr,pandya2015icra,km2017icra,wang20206_6pack}.
However, canonicalization approaches require painstakingly gathered 3D data for supervision~\cite{nocs,xnocs} or only provide \emph{sparse} keypoint-based representations~\cite{c3dpo}, making them impractical for real-world operation.

\begin{figure}[ht!]
    \centering
    \includegraphics[width=0.9\columnwidth]{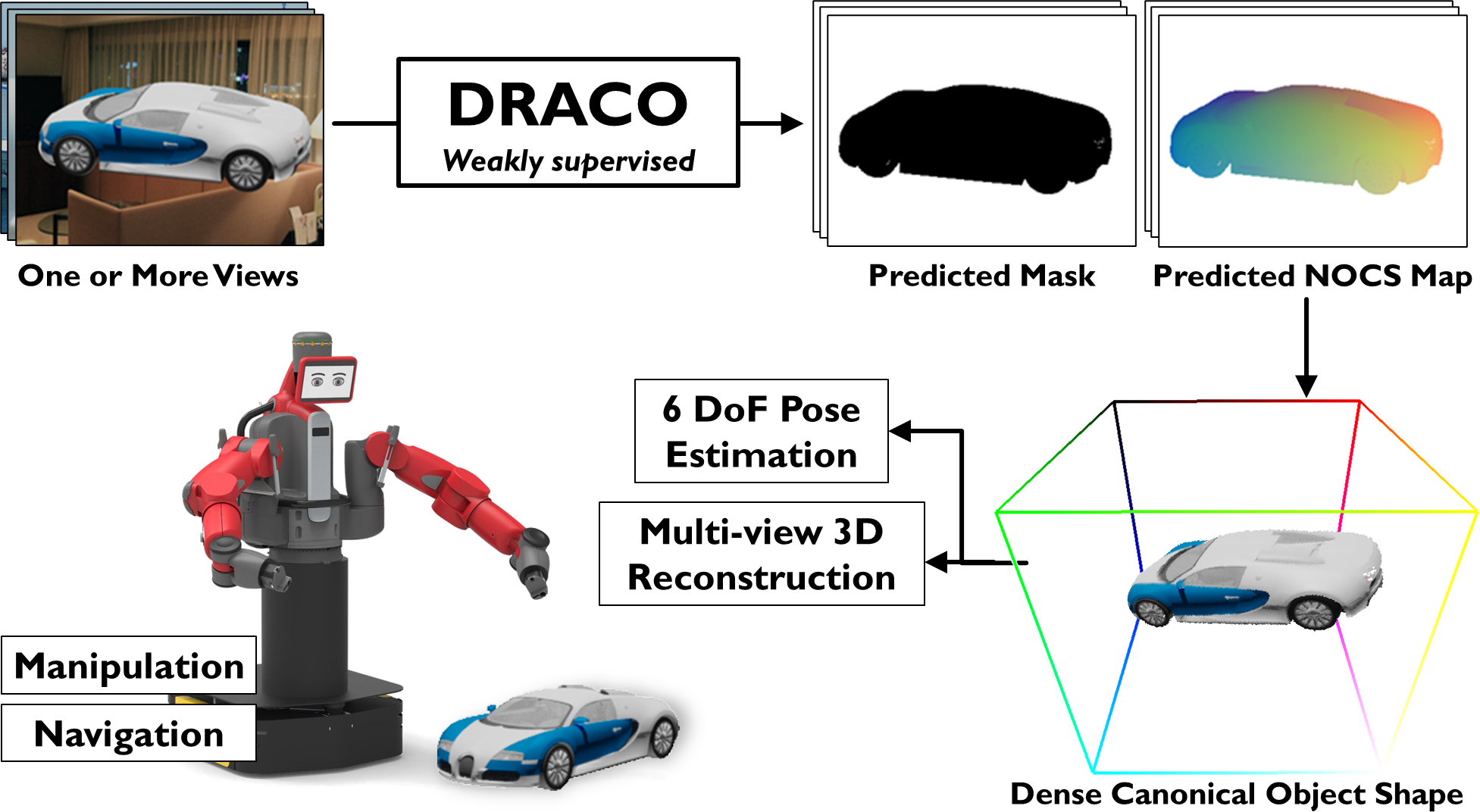}
    \vspace{-0.05in}
    \caption{\textbf{\coolname} is a category-level reconstruction method that performs \emph{dense canonicalization}. This is useful, for instance, in robot manipulation of a toy car with estimated shape and 6~DoF pose.
    We use NOCS maps~\cite{nocs} as the canonical shape representation.
    \label{fig:teaser}
    \vspace{-0.15in}
    }
\end{figure}

We address this challenge with DRACO, a \emph{weakly supervised} method for \underline{D}ense \underline{R}econstruction \underline{A}nd \underline{C}anonicalization of \underline{O}bject shape from one or more images (Fig.~\ref{fig:teaser}).
We estimate the 3D shape of an object in a coordinate system canonicalized for scale and rototranslation parameters without using any dense supervision in 3D.
We adopt \emph{NOCS maps}~\cite{nocs} to represent the canonical shape of visible object surfaces (see Fig.~\ref{fig:teaser}, \ref{fig:nocs}).
Different from previous work~\cite{c3dpo,nocs}, \textbf{we estimate dense NOCS maps without using dense supervision}.
Our approach has two main contributions: (1)~a weakly supervised \emph{object-centric} depth estimation network, and (2)~an unsupervised keypoint-guided dense canonicalization method.

We design a CNN to estimate depth maps using easily acquired weak-supervision (foreground object masks and camera motion).
Different from prior work~\cite{godard2017unsupervised}, we build \emph{object-centric} depth maps that only reason about visible object pixels.
By enforcing multi-view photometric and geometric consistency, accurate segmentation, and smoothness, we produce high-quality object-centric depth maps.
In isolation, these depth maps cannot be used in downstream tasks due to the absense of canonicalization.
Therefore, we learn to \emph{lift} and canonicalize sparse 2D keypoints to 3D with C3DPO~\cite{c3dpo}.
These canonical 3D keypoints and object-centric depth maps then aid the unsupervised learning of NOCS maps.
At inference time, \coolname performs \emph{dense canonicalization} of object shape and pose for \textbf{previously unseen} instances from one or more RGB images \textbf{without any additional supervision}.
Our main contributions include:
\begin{itemize}
    \item weakly-supervised \emph{object-centric} depth prediction leveraging a new consistency-seeking perceptual loss, and
    \item unsupervised dense canonicalization to estimate NOCS maps by maximizing geometric consistency.
\end{itemize}

Extensive experiments on our new large-scale mixed reality dataset (20K images/category) demonstrate superior shape reconstruction and canonicalization quality compared to state-of-the-art fully supervised approaches~\cite{xnocs}, and weakly-supervised baselines that strictly use more information.
Although trained only with mixed reality images, \coolname is able to generalize to real images at inference time (see Fig.~\ref{fig:real_results}).
Dense canonicalization also enables trivial aggregation of multi-view shape cues via a set union operation.
\textbf{All code, data, and models will be made public}.

\begin{figure*}[!t]
    \centering
    \includegraphics[width=0.95\textwidth]{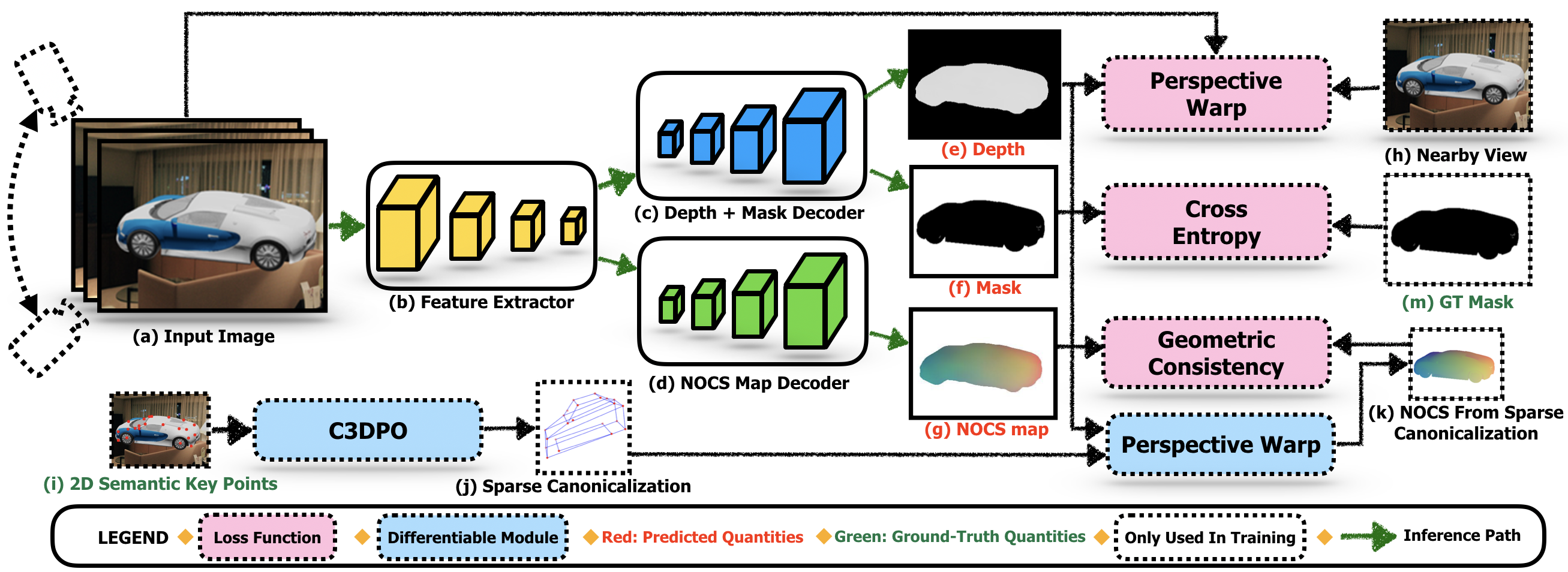}
    \caption{\textbf{System overview}: One or more input RGB images (a) are passed through the feature extractor (b) and the depth, mask, and NOCS map decoders (c, d). The output depth (e) and mask (f) are used to compute an inverse perspective warp that synthesizes view (a) from nearby view (h) to weakly-supervise object-centric depth prediction. At train time, we leverage camera motion and 2D semantic keypoints (i) to initiate a sparse canonicalization based on C3DPO~\cite{c3dpo}.
    The sparse canonicalization (j) is densified using the predicted object-centric depth (e), which is warped to match (g). This discrepancy defines the \emph{geometric consistency} loss used in the unsupervised learning of NOCS map.
    At inference time, \coolname only requires a single image to predict object-centric depth (e) and NOCS map (g).
    }
    \label{fig:pipeline}
    \vspace{-0.2cm}
\end{figure*}
\raggedbottom

\section{Related Work}

Our work broadly relates to a number of themes in the robotics and computer vision communities.

\parahead{Dense Correspondence Estimation}
Many robotics problems that involve object or scene level understanding require dense correspondences of pixels across a pair of images. Examples include dense simultaneous localization and mapping~\cite{densevo, kinectfusion}, manipulation~\cite{florence2018dense}, and non-rigid tracking~\cite{newcombe2015dynamicfusion}. Dense canonicalization can be perceived as a stricter subset of dense correspondence estimation, as it involves not just associating corresponding pixels across views--rather mapping pixels to a category-level canonical coordinate frame.

\parahead{Unsupervised Depth Estimation}
Several approaches have focused on recovering dense depth measurements from a single image (\eg,~\cite{garg2016unsupervised,zhou2017unsupervised,monodepth2,packnet}). Typically, such approaches leverage the photometric multi-view consistency of a static scene at train time. However, such single-view depth estimation approaches provide camera-centric depth estimates for the \emph{entire image}, while the idea of canonicalization is to produce \emph{object-centric} canonical 3D coordinates. 

\parahead{Category-Level Perception}
%
Category-level perception is important in robotics applications that require reasoning in unconstrained environments~\cite{manipulationbook,kemp2007challenges,edsinger2007robot,campbell2010autonomous, thrun2006stanley, dsdnet,goodrich2008human}. This is evident from the immense interest in 6D pose estimation~\cite{wang20206_6pack,chen2020learning,manhardt2020cps,nocs}, object-based SLAM~\cite{burchfiel2017bayesian,parkhiya2018icra,mu2016iros,nicholson2018quadricslam,mccormac2018fusion++},articulated category-level pose estimation~\cite{li2020category,kulkarni2020articulation}, and autolabeling~\cite{zakharov2020autolabeling}.
The notion of canonicalization has been applied to practical robotic tasks such as rigid-body manipulation~\cite{manuelli2019kpam}, visual servoing~\cite{pandya2015icra}, and multi-object tracking~\cite{beyondpixels}.
Canonical surface mapping (CSM)~\cite{kulkarni2019canonical} is a recent approach that learns to map every instance of an object category to a \emph{canonical instance} or prototype. This restricts the expressability of the class of shapes that can be represented in a CSM. In particular, CSMs necessitate fixed topology and correspondences across instances, whereas our approach---building on NOCS~\cite{nocs}---is a category-level representation agnostic to intra-category variations in topology.
Current approaches leveraging canonicalization such as X-NOCS~\cite{xnocs} leverage dense, aligned 3D scans of an object category for multi-view shape reconstruction~\cite{xnocs}. In a similar vein, category-level canonicalization has been applied to surface reconstruction~\cite{lei2020pix2surf}.
In contrast to prevailing techniques, \coolname does not require dense 3D supervision, and exhibits better generalization to real images.


\section{Background}
\begin{figure}[H]
    \centering
    \includegraphics[width=0.95\columnwidth]{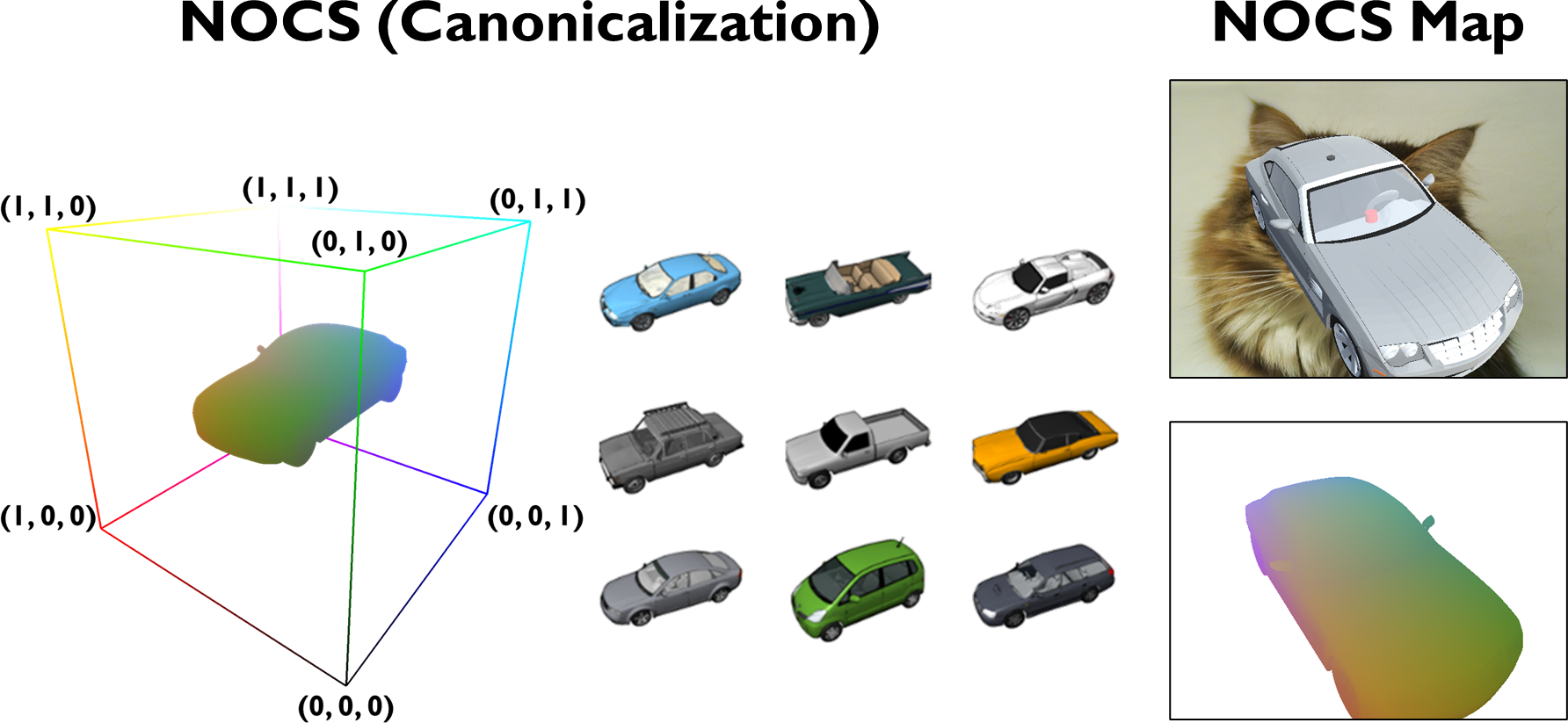}
    \caption{
    \textbf{NOCS} (Normalized Object Coordinate Space) is a category-level canonical coordinate frame (a unit cube). Pixel intensities in the NOCS map determine its 3D location in the NOCS cube.
    }
    \label{fig:nocs}
\end{figure}

\parahead{Normalized Object Coordinate Space (NOCS)}
NOCS~\cite{nocs} is a category-specific normalized 3D canonical space that can contain different object instances. A NOCS map is a color-coded 2D projection of the object's 3D NOCS shape -- a pixel in the NOCS map with an RGB color value (range: 0--1)
will be located at the corresponding normalized position $(x, y, z)$, $\forall$ $x, y, z \in [0, 1]$ (see Fig.~\ref{fig:nocs}). NOCS is a \textbf{canonical frame} -- corresponding parts of different instances of an object category are located at a similar location in the NOCS container.

\parahead{C3DPO}
NOCS can be used for perception tasks like pose estimation~\cite{nocs,li2020category}, 3D reconstruction~\cite{xnocs,lei2020pix2surf} and autolabeling~\cite{zakharov2020autolabeling}, although it requires large amounts of supervision.
Recently, C3DPO~\cite{c3dpo} proposed a method for \emph{unsupervised} learning of a canonical frame given semantic 2D object keypoints and their visibility. Given keypoints in multiple views, a factorization network estimates a transformation matrix $T_{\text{C3DPO}}$, shape basis vectors $S_{\text{C3DPO}}$, and their view specific pose parameters $\alpha_{\text{C3DPO}}$. It captures the view invariant shape $X_{\text{C3DPO}}$ that is aligned to the camera frame by $T_{\text{C3DPO}}$. 
Our method builds upon the transformation obtained by C3DPO and uses it as a prior while learning the dense canonicalized NOCS map for an object.




\section{DRACO: Dense reconstruction and canonicalization of objects}
\label{sec:draco}


Given an image $I$ of an object from a prescribed category, we seek to learn a \emph{dense 3D canonicalization map} $f_\theta$ of that object. This map $f_\theta$ transforms every image pixel belonging to the object to a unique 3D location in the NOCS container. This canonicalization is invariant to scale, rotation, and translation across all instances of the category.

\textbf{Our goal is to learn this dense canonicalization $f_\theta$ without requiring dense 3D supervision} (\ie,~in the form of ground truth NOCS maps or depth maps), which are painstaking to gather. At training time, we assume that each input image with intrinsics $K$, and two nearby views $I_1$ and $I_2$ with extrinsics $T_1$ and $T_2$ respectively. To capture category-level information, we assume a sparse set of 2D \emph{keypoints} and a binary foreground mask $M$ delineating the object. At inference time, the learned dense canonicalization operates on a single un-posed input image but \textbf{without any additional input} such as foreground masks or keypoints.


\subsection{Overview}
Weakly supervised dense canonicalization is an extremely challenging problem due to several reasons. First, it requires a 3D understanding of the relative depth of each image pixel corresponding to the object. Second, it necessitates a mapping from the camera coordinate frame to the canonical frame that is consistent across all instances of the category.
%
We address these challenges by building a dense canonicalization function $f_\theta(I)$ as a convolutional neural network (CNN) that takes as input an image $I$ and predicts a NOCS map $\hat{N}$.
Prior approaches such as X-NOCS~\cite{xnocs} learn $f_\theta$ by supervised learning against ground truth NOCS maps. In contrast, we do not assume the availability of ground truth NOCS maps, and instead propose to learn depth prediction and sparse canonicalization in a weakly supervised fashion.

Our overall architecture comprises of 3 components: a feature extractor (encoder), a depth and mask decoder, and a NOCS map decoder (see Fig.~\ref{fig:pipeline}). We encode the image intensity information into a feature space using a fully convolutional feature extractor (ResNet-50~\cite{resnet}). A depth decoder and an object mask predictor operate over these encoded features to produce a dense 3D reconstruction in the camera coordinate frame. For the canonicalization of these depth maps, we propose a keypoint-guided method that builds upon C3DPO~\cite{c3dpo} to canonicalize a sparse set of semantic keypoints. Given a set of keypoints and their visibilities, C3DPO produces a sparse canonicalization transform that best aligns semantic keypoints across all instances of the category. In a parallel branch to the depth and mask decoders, we train a NOCS map decoder that generates a NOCS map $\hat{N}$, solely relying on the encoded features.

The predicted depth maps $\hat{D}$ and the sparse canonicalization are used to generate an independent estimate of the NOCS map $\hat{N}_{depth}$. If the predicted depths, masks, canonicalizations, and NOCS maps are all accurate, we expect $\hat{N} = \hat{N}_{depth}$. In practice, this is not the case, and hence the gradient (w.r.t.~the parameters $\theta$) of $\|\hat{N} - \hat{N}_{depth}\|$ can be used as a loss function to train $f_\theta$, without ever requiring access to the ground-truth NOCS maps or 3D annotations of any form.
Importantly, \textbf{we require only a single image $I$ at inference time}. We first train the depth and mask decoders, after which we proceed to train the NOCS map decoder. We find this stabilizes training and improves convergence.



\subsection{Architecture and Loss Functions}


\begin{figure*}[!tbh]
\centering
\vspace{2mm}
\includegraphics[width=0.98\linewidth, height=130pt]{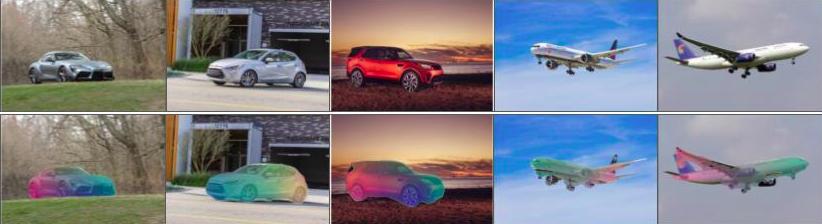}
\caption{\textbf{Real-world results}: We train \coolname{} using the mixed reality dataset and test it (without any finetuning) on real-world images (obtained by scraping Google images for query keywords \texttt{car} and \texttt{airplane}). \coolname{} (bottom row) generalizes remarkably well to this out-of-training-distribution data. We attribute this performance to the color jittering performed by \coolname{} during training. Also, note how corresponding parts across instances (the nose and tail of airplanes and the front and rear of cars) get accurately canonicalized (similar parts have similar NOCS map colors).
}

\label{fig:real_results}
\end{figure*}

\parahead{Object-centric depth estimation}
Our depth decoder builds on prior work~\cite{monodepth2, sfmlearner} that leverages geometric consistency for learning depth from a single image. The decoder has 4 upconvolution layers, with each block comprising two convolution layers with an exponential linear unit~\cite{elu} and an instance normalization layer~\cite{instancenorm}. Skip connections between the encoder and decoder layers are applied to preserve fine-grained image details, often useful in depth prediction.
In line with~\cite{monodepth2, sfmlearner, ssim}, we use a photometric consistency loss in conjunction with regularizers that enforce smoothness. Following~\cite{Johnson2016Perceptual}, we also employ a perceptual loss to preserve high-frequency details. The photometric loss employs an inverse perspective warp $w(I_i, \hat{D}, K, T_i)$ to synthesize the target view $\hat{I}$ from a nearby view $I_i$, followed by a structural similarity metric (SSIM~\cite{ssim}) to compare it with the target view $I$, where $i \in \{1, 2\}$. 
\begin{equation}
    \label{eqn:photo}
    \mathcal{L}_{\text{ph}} = \left( \frac{\alpha}{2} (1 - \text{ssim}(\hat{I}, I)) + (1 - \alpha) \| \hat{I} - I \|_1 \right) \odot M
\end{equation}
Here, $\alpha \in \left[0, 1\right]$ is a hyperparameter that weighs high-frequency details against an L1 image difference. We empirically set $\alpha$ to $0.15$ in all our experiments.

To encourage smoother depth predictions at low-frequency regions in the image, we use a regularization penalty. $\delta_{x}$ and $\delta_{y}$ denote the image gradients along the $x$ and $y$ directions,
\begin{equation}
    \mathcal{L}_{\text{smooth}} = (|\delta_{x} \hat{D}|e^{-|\delta_{x} I|} + |\delta_{y} \hat{D}|e^{-|\delta_{y} I|}) \odot M
\end{equation}

We additionally employ a perceptual loss to obtain better quality depth estimates, following~\cite{Johnson2016Perceptual}. Using a VGG-16~\cite{vgg} network pretrained on ImageNet, we extract feature maps from the \texttt{relu2\_2} layer and zero out all feature responses outside the receptive field of the foreground mask $M$. Denoting this operation $\text{feat}(., .)$, the perceptual loss is
\begin{equation}
    \mathcal{L}_{\text{per}} = \| \text{feat}(I, M) - \text{feat}(\hat{I}, M) \|_{1}
\end{equation}

\parahead{Mask Decoder}
The mask decoder has a similar architecture as the depth decoder, ending in a sigmoid activation function. This decoder predicts a foreground mask $\hat{M}$ and is supervised by the ground-truth mask $M$, using a binary cross entropy loss. 
\begin{equation}
    \mathcal{L}_{\text{m}} = -\mathbb{E}_{i \in M} \left[M_i \log{\hat{M}_i} + (1 - M_i) \log{(1 - \hat{M}_i)} \right]
\end{equation}

\parahead{Keypoint-guided canonicalization}
\label{sec:nocs_from_depth}
At training time, we leverage category-level semantic keypoints (2D) $kp$ to estimate a \emph{sparse canonicalization} of the object. We apply the factorization network $\Phi$ from C3DPO~\cite{c3dpo} to recover the canonical orientation of the object. The translation vector and scale factor can be recovered by aligning this sparse point cloud with the image keypoints backprojected to 3D via the predicted depth map $\hat{D}$. We denote this canonicalization transform by $T_{\text{cano}}$.
In conjunction with the predicted depth $\hat{D}$, this transform $T_{\text{cano}}$ can be used to produce a NOCS representation $\hat{N}_{depth}$ of the object of interest: we backproject image points within the mask $M$ using the estimated depth $\hat{D}$, and apply a rigid transform $T_{\text{cano}}$ to canonicalize this dense pointcloud (\cf. lower C3DPO branch, Fig. \ref{fig:pipeline}).

\parahead{NOCS Map Decoder}
In a parallel branch to the depth and mask decoders, a NOCS map decoder predicts a NOCS map corresponding to the central object in the image. This decoder is similar to the depth and mask decoders. Since we do not assume the availability of ground-truth NOCS maps, we instead impose a photometric and geometric consistency criterion to train this decoder. The key insight here is that once canonicalized, corresponding points across multiple views must map to the same 3D location inside the NOCS container. We apply an inverse perspective warp to the predicted NOCS map $\hat{N}_i$ for view $i$ and enforce equality with the predicted NOCS map $\hat{N}$ of the target view.
\begin{equation}
    \label{eqn:nocs_depth_consistent_loss}
    \mathcal{L}_{\text{geo}} = \| w(\hat{N}_i, \hat{D}, K, T_i) - \hat{N} \|_1 \odot M
\end{equation}
Similar to the depth decoder, we also compute perceptual loss $\mathcal{L}_{\text{per}}$ and photometric loss $\mathcal{L}_{\text{ph}}$ between $\hat{N}$ and the independent NOCS estimate $\hat{N}_{depth}$ (Sec. \ref{sec:nocs_from_depth}) and a smoothness regularizer $\mathcal{L}_{\text{smooth}}$ over the predicted NOCS map $\hat{N}$.

\parahead{Loss Functions}
The aggregate loss functions for the depth and nocs decoders are the following.
\begin{equation}
    \mathcal{L}_{\text{depth}} = \mathcal{L}_{\text{ph}} + \mathcal{L}_{\text{smooth}} + \mathcal{L}_{\text{per}}
\end{equation}
\begin{equation}
    \mathcal{L}_{\text{nocs}} =  \mathcal{L}_{\text{geo}}   +
    \mathcal{L}_{\text{ph}} +
    \mathcal{L}_{\text{per}}   + \mathcal{L}_{\text{smooth}}
\end{equation}

\parahead{Training}
We first train the depth and mask decoders for $20$ epochs using $\mathcal{L}_{\text{depth}}$ and $\mathcal{L}_{\text{m}}$, after which we simultaneously begin training the NOCS map decoder using $\mathcal{L}_{\text{nocs}}$. We only use gradients with respect to $\mathcal{L}_{\text{nocs}}$ to update the parameters of the NOCS map decoder, to prevent trivial solutions where depth and NOCS maps shortcut to zeros. We augment training images by color space jitter and train for a total of 40 epochs.
In all our experiments, we use the Adam~\cite{adam} optimizer with a learning rate of $5e^{-6}$. At epochs 10, 15, and 18, we decay the learning rate by a factor of 0.1. We downsample input images to a resolution of $640 \times 480$.

\section{Experiments and Results}
We present an extensive experimental evaluation of design choices and compare \coolname with other supervised and weakly supervised canonicalization approaches.
Unlike previous work~\cite{xnocs,nocs}, we study a variety of parameters, including separate experiments on shape quality, canonicalization accuracy, and pose estimation accuracy.

\begin{figure}
\vspace{1.2mm}
\centering
\includegraphics[height=114pt,width=\linewidth]{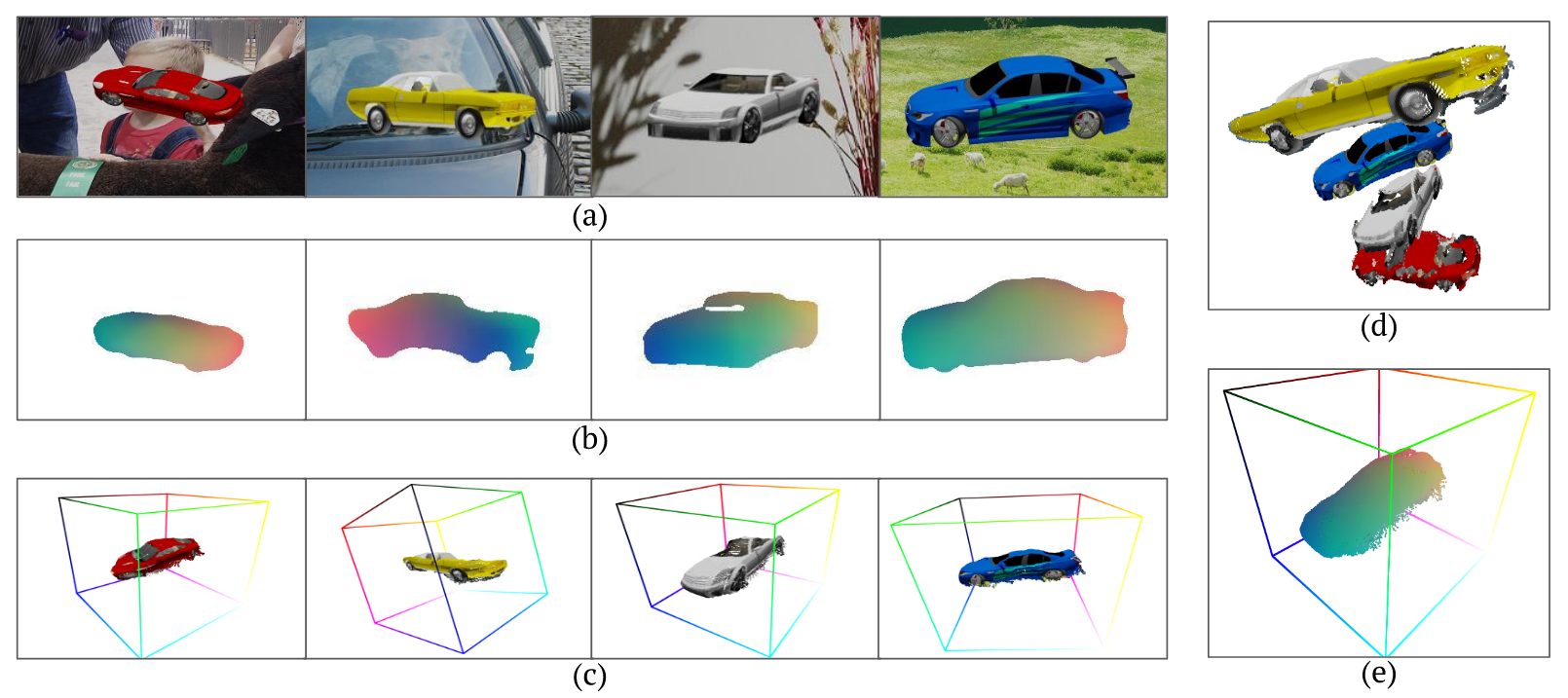}
\caption{
\textbf{Qualitative results} illustrating pose and shape canonicalization: (a) input mixed reality images, (b) predicted NOCS maps, (c) recovered 3D shape from NOCS maps, (d) reconstructed shapes without canonicalization (visualized with varying vertical displacements), (e) all reconstructed objects overlaid in the NOCS cube. Notice how the various instances align together and exhibit a consistent scale and rototranslation.
}
\label{fig:syn_results}
\end{figure}

\parahead{Dataset}
\label{sec:dataset}
There are no well-established real datasets for evaluating shape reconstruction and canonicalization due to the difficulty in obtaining ground truth labels.
Most prior works resort to synthetic or \emph{mixed reality} data~\cite{nocs}, but these datasets lack keypoint annotations, which we need for training.
We, therefore, generate our own mixed reality dataset, \textbf{DRACO20K}, which consists of foreground images rendered using synthetic objects from ShapeNet~\cite{chang2015shapenet}, augmented with background images from MS COCO~\cite{lin2015microsoft}.
For each instance, we render 50 views in a helical path around the object for a total of 20K images per category.
We vary the specular, roughness, and metallic parameters of instances to emulate real-world reflections and reduce the sim2real gap as shown in Fig.~\ref{fig:syn_results}.
We also store the ground truth depth maps, NOCS maps (not used in training), object mask, camera pose, keypoints~\cite{you2020keypointnet}, and keypoint visibility.

\raggedbottom

\parahead{Metrics}
\label{sec:comp_metrics}
We quantify our reconstruction and canonicalization quality using several metrics. The \textbf{reconstruction quality} of predictions is compared using a 2-way Chamfer distance (3D loss) and an L1 photometric loss (2D loss) after instance-wise alignment of the predicted canonical maps to the ground truth NOCS maps.
Since we use C3DPO~\cite{c3dpo} for keypoint canonicalization, our canonical frame of reference is different from the ground truth NOCS canonical frame.
We compute an optimal average transformation between our predicted canonical frame and the NOCS frame by performing instance-wise Umeyama alignment~\cite{umeyama1991least}, and ICP followed by RANSAC for the entire dataset.
We report the \textbf{reconstruction + canonicalization quality}~(see Sec.~\ref{sec:nocs_cmp}) by computing errors both in 3D (Chamfer) and in 2D ($L1$ photometric error) between the ground truth NOCS maps and our predicted maps given this optimal average transform.
We also evaluate performance on 6~DoF pose and size estimation tasks~\cite{nocs} (Sec. \ref{sec:pose_est}) using standard metrics, and empirically analyze the impact of canonicalization by comparing the pre- and post-canonicalization distributions (Sec.~\ref{sec:can_hist}).

\subsection{Baselines and Ablations}
\label{sec:comp_baselines}
To investigate the impact of our design choices, we implement two weakly supervised baselines that use strictly more information (semantic keypoint) at inference time.
\textbf{Baseline-kps} predicts the depth maps of input RGB images using an ResNet-50 encoder-decoder.
The predicted depth maps are then canonicalized using ground-truth keypoints fed to C3DPO~\cite{c3dpo} that outputs the \emph{best-possible} learned sparse canonicalization.
\textbf{Baseline-kps-noisy} is similar to Baseline-kps, except that it uses keypoint predicted from a stacked hourglass network~\cite{shape_priors_kp_net} in place of ground-truth, to emulate real-world settings where the keypoints are often noisy or incorrect.
We also compare against an \textbf{Oracle} that canonicalizes ground-truth depth maps using the optimal transform.




Table~\ref{table:NOCS_comparison} compares the different baselines with our approach. Our method achieves superior 3D reconstruction quality evident from consistently lower Chamfer and L1 NOCS errors when compared to Baselines-kps and Baseline-kps-noisy. Notably, \coolname{} requires only an RGB image information at inference time.
Baseline-kps-noisy uses predicted keypoints that degrades its performance due to inaccurate keypoint locations and visibility. Imprecise keypoint predictions give erroneous scale and transformation while performing Umeyama alignment (Sec. \ref{sec:nocs_from_depth}). For Baseline-kps, although we use ground truth keypoints, we observe a deterioration in its quality of canonicalization in Tab. \ref{table:NOCS_comparison} and Fig. \ref{fig:map}.
Baseline-kps is unable to recover from inaccuracies introduced by C3DPO, explaining this performance drop.
However, this is not the case for \coolname, because our geometric consistency loss (Eqn. \ref{eqn:nocs_depth_consistent_loss}) enables recovery from noisy sparse canonicalization.
Our method does not need keypoints during inference and instead relies on the predicted depth and multiview consisteny to refine NOCS maps.

\begin{table}[t]
\vspace{2.5mm}
\centering
\resizebox{.45\textwidth}{!}{
\begin{tabular}{c|c|c|c|c|c}
\toprule
\multicolumn{2}{c|}{} &
  \multicolumn{2}{c|}{{\color[HTML]{000000} \textbf{Reconstruction}}} &
  \multicolumn{2}{c}{{\color[HTML]{000000} \textbf{\begin{tabular}[c]{@{}c@{}}Reconstruction +\\  Canonicalization\end{tabular}}}} \\ \cline{3-6} 
\multicolumn{2}{c|}{\multirow{-2}{*}{\textbf{Model}}} &
  {\color[HTML]{000000} \textbf{\begin{tabular}[c]{@{}c@{}}Chamfer $\downarrow$\end{tabular}}} &
  {\color[HTML]{000000} \textbf{\begin{tabular}[c]{@{}c@{}}L1 $\downarrow$ \end{tabular}}} &
  {\color[HTML]{000000} \textbf{\begin{tabular}[c]{@{}c@{}}Chamfer $\downarrow$\end{tabular}}} &
  {\color[HTML]{000000} \textbf{\begin{tabular}[c]{@{}c@{}}L1 $\downarrow$\end{tabular}}} \\ \midrule
 &
  \cellcolor[HTML]{FFE3E3}\textbf{Oracle} &
  \cellcolor[HTML]{FFE3E3}-- &
  \cellcolor[HTML]{FFE3E3}-- &
  \cellcolor[HTML]{FFE3E3}4.635 &
  \cellcolor[HTML]{FFE3E3}1.388 \\  
 &
  \textbf{Baseline-kps~\cite{c3dpo}} &
  2.832 &
  2.106 &
  8.607 &
  2.742 \\  
 &
  \textbf{Baseline-kps-noisy\cite{c3dpo}+\cite{shape_priors_kp_net}} &
  8.911 &
  2.471 &
  14.451 &
  3.108 \\  
 &
  \textbf{\coolname (Ours)} &
  \textbf{2.099} &
  \textbf{2.484} &
  \textbf{5.183} &
  \textbf{2.487} \\  
\multirow{-5}{*}{\textbf{\rotatebox[origin=c]{90}{Car}}} &
  \cellcolor[HTML]{ECF4FF}\textbf{XNOCS (Sup.)~\cite{xnocs}} &
  \cellcolor[HTML]{ECF4FF}3.272 &
  \cellcolor[HTML]{ECF4FF}2.004 &
  \cellcolor[HTML]{ECF4FF}3.575 &
  \cellcolor[HTML]{ECF4FF}2.032 \\ \midrule
 &
  \cellcolor[HTML]{FFE3E3}\textbf{Oracle} &
  \cellcolor[HTML]{FFE3E3}-- &
  \cellcolor[HTML]{FFE3E3}-- &
  \cellcolor[HTML]{FFE3E3}7.098 &
  \cellcolor[HTML]{FFE3E3}1.133 \\  
 &
  \textbf{Baseline-kps~\cite{c3dpo}} &
  8.457 &
  3.712 &
  17.676 &
  3.805 \\  
 &
  \textbf{Baseline-kps-noisy~\cite{c3dpo}+\cite{shape_priors_kp_net}} &
  19.716 &
  4.946 &
  48.201 &
  5.157 \\  
 &
  \textbf{\coolname (Ours)} &
  \textbf{4.192} &
  \textbf{2.981} &
  \textbf{8.803} &
  \textbf{3.055} \\  
\multirow{-5}{*}{\textbf{\rotatebox[origin=c]{90}{Airplane}}} &
  \cellcolor[HTML]{ECF4FF}\textbf{XNOCS (Sup.)~\cite{xnocs}} &
  \cellcolor[HTML]{ECF4FF}18.123 &
  \cellcolor[HTML]{ECF4FF}2.521 &
  \cellcolor[HTML]{ECF4FF}18.191 &
  \cellcolor[HTML]{ECF4FF}2.623 \\ \bottomrule
\end{tabular}}
\caption{\textbf{Reconstruction and canonicalization errors} of \coolname{} compared to X-NOCS and other strong baselines. We obtain performance competitive to X-NOCS; a fully supervised approach. We are strictly better than baselines \texttt{kps} and \texttt{kps-noisy} which use privileged information (semantic keypoints) \emph{at test time}. We report the Chamfer distance between the 3D NOCS pointclouds and the $L1$ difference (2D) between the estimated and true NOCS maps (multiplied by $10^{3}$). \coolname only requires an RGB input image at inference time.}
\label{table:NOCS_comparison}
\end{table}
\raggedbottom

\begin{table}[H]
\centering
\scalebox{0.9}{
\begin{tabular}{l|c|c}
\toprule
\textbf{\begin{tabular}[c]{@{}c@{}}Training losses\end{tabular}} &
  \multicolumn{1}{c|}{\textbf{\begin{tabular}[c]{@{}l@{}}Photometric error $\downarrow$ \end{tabular}}} &
  \multicolumn{1}{c}{\textbf{\begin{tabular}[c]{@{}l@{}}L1 (depth) $\downarrow$  \end{tabular}}} \\ \midrule
L1 + SSIM                                                                   & 0.1317          & 0.147         \\
OURS (L1 + SSIM + Perceptual)                                             & \textbf{0.1311} & \textbf{0.143} \\
L1 + SSIM + Chamfer                                                         & 0.1313          & 0.15           \\
\begin{tabular}[c]{@{}l@{}}L1 + SSIM + Perceptual + \\ Chamfer\end{tabular} & 0.1315          & 0.149          \\ \bottomrule
\end{tabular}
}
\caption{\textbf{Ablation analysis} of various loss function combinations (lower photometric and L1 (depth) errors are better).}
\label{table:depth_ablation}
\end{table}
\raggedbottom

\parahead{Depth Estimation Loss Variants}
We perform an ablation analysis on the choice of loss functions for object-centric depth estimation.
Table \ref{table:depth_ablation} compares the performance of the different loss functions. For each loss function considered, we report (1)~the L1 (depth) error, which is the mean L1 distance between the predicted depth and ground truth depth map within the mask, and (2)~the photometric loss $\mathcal{L}_{ph}$ (see Eqn.~\ref{eqn:photo}). Note that the ground-truth depth maps are only used for error evaluation, and are not available at train time.
We observe that our final loss function performs the best, and note that adding a Chamfer loss qualitatively degrades depth and introduces smearing artifacts.

\subsection{Reconstruction and Canonicalization}
\label{sec:nocs_cmp}
We also compare our reconstruction and canonicalization quality with X-NOCS~\cite{xnocs}, a state-of-the-art fully supervised dense canonicalization approach.
Dense NOCS map supervision is hard to obtain in real images, while our approach only uses easy-to-acquire weak masks and pose labels.
\definecolor{bluerow}{HTML}{ECF4FF}
\sethlcolor{bluerow}
The \hl{\mbox{highlighted row}} for each category in Tab.~\ref{table:NOCS_comparison}  shows quantitative comparison with X-NOCS.
We observe that our approach performs better in shape reconstruction and comparable in reconstruction+canonicalization quality compared to X-NOCS.
Our method constrains the neighboring pixels using the smoothness loss (absent in X-NOCS), which results in better shape reconstruction of the canonicalized object.
Moreover, using perceptual loss enables us to capture intricate object details better than using L2 loss alone.
We also perform on par with X-NOCS in the L1 NOCS map loss.
This shows that despite using weak supervision, we are able to estimate high-quality shape and canonicalization (see Fig. [\ref{fig:real_results}, \ref{fig:syn_results}] for visual results).

\begin{figure}[!tbh]
\centering
\includegraphics[width=0.5\textwidth]{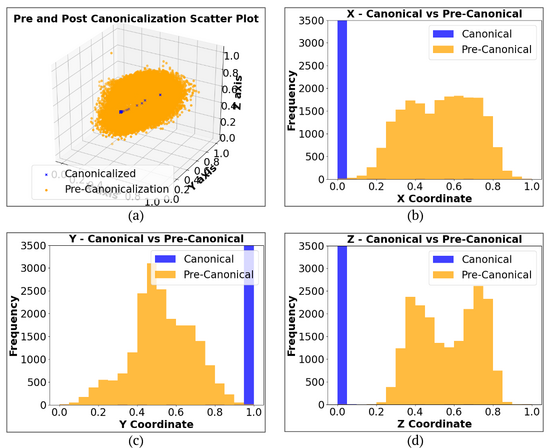}
\caption{\textbf{Impact of canonicalization}: We plot the distribution of a keypoint (rear-left bumper) in 3D (a) and along the X, Y, and Z axes respectively (b-d). When devoid of canonicalization (yellow), the keypoint distribution has significantly greater entropy (note the dispersion in the scatter plot and the histograms). After canonicalization (blue), the keypoint distribution is a single peak, which demonstrates the impact of our canonicalization scheme.
}
\label{fig:canon_hist}
\end{figure}

\parahead{Canonicalization Histograms}
\label{sec:can_hist}
To evaluate canonicalization quality decoupled from reconstruction, we plot the distribution of 3D keypoint positions, pre- and post-canonicalization, as scatter plots and histograms in Fig.~\ref{fig:canon_hist}.
As seen in the figure, the objects' keypoints are heavily scattered in the normalized $(x, y, z)$ coordinate space before canonicalization.
However, after canonicalization, the keypoints are transformed into certain fixed localities of the normalized coordinate space which is evident by the sharp peak and in the concentrated cluster of keypoints post canonicalization in Fig. \ref{fig:canon_hist}.
This shows that our canonicalizations are consistent and unique.

\subsection{6D Pose and Scale Estimation}
\label{sec:pose_est}
Canonical shape reconstruction is useful in many downstream perception tasks like 6~DoF pose estimation critical for robot manipulation and navigation.
We evaluate DRACO in the task of 6~DoF pose and scale estimation.
We follow the protocol and alignment method of NOCS~\cite{nocs} to estimate the 6~DoF pose and scale parameters using Umeyama's algorithm~\cite{umeyama1991least}.
We compare the pose and scale from the NOCS map generated by our method as well as X-NOCS. 

\begin{figure}[!tbh]
\centering
\includegraphics[width=0.4\textwidth]{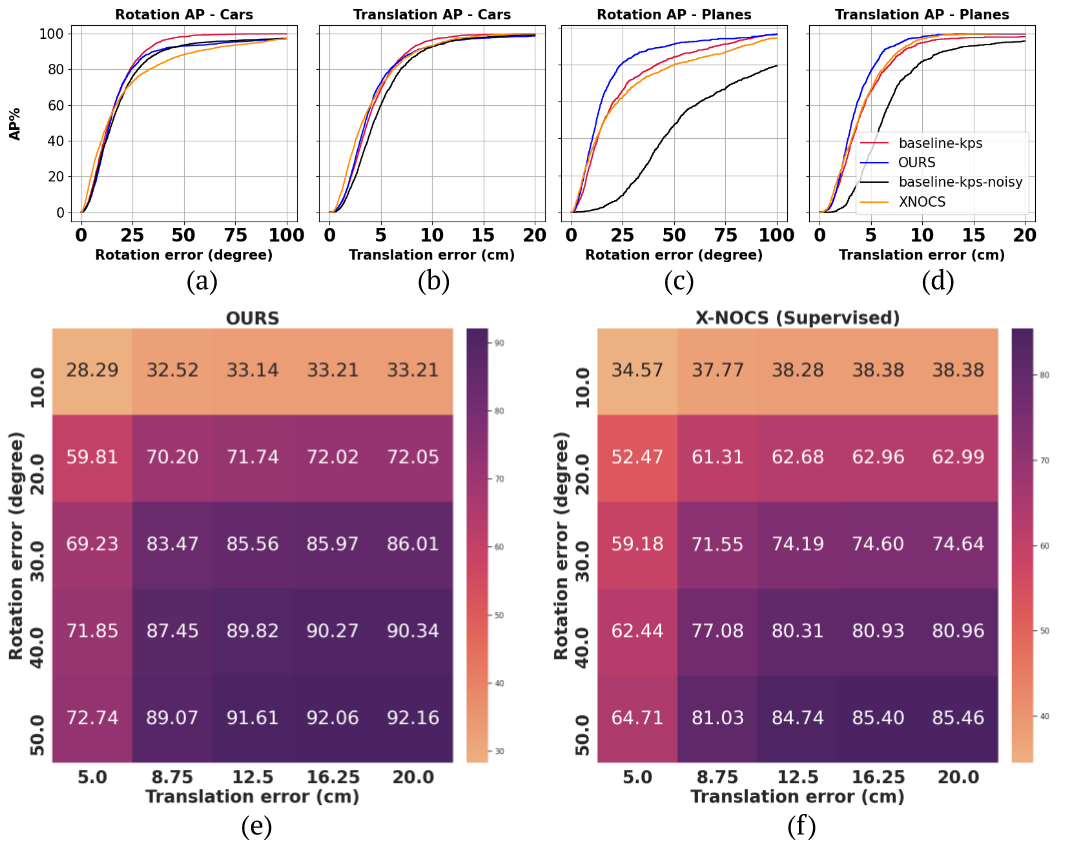}
\caption{\textbf{Pose estimation accurary}: We report the mean average precision (mAP) for rotations and translations independently for the \texttt{car} and \texttt{plane} classes (a-d).
To evaluate 6-DoF pose estimation accuracy, we also plot heatmaps of \coolname (OURS) and X-NOCS. Our coarse-grained performance is superior to X-NOCS: we achieve a mAP of $85.97\%$ in the $(30\degree, 15\mbox{cm})$ metric.
}
\label{fig:map}
\end{figure}

Fig. \ref{fig:map} reports the mean average precision (mAP) curves of rotation and translation error in 7D pose estimation for \coolname, X-NOCS, and all the baselines. The heatmaps represent the percentage of data within the defined rotation and translation errors for DRACO and X-NOCS.
As seen from Figure \ref{fig:map}, we attain a mAP of $85.97\%$ in the $(30\degree, 15 \mbox{cm})$ metric despite not using dense supervision, which is higher than X-NOCS $(74.54\%)$.
Our mAP results are better than X-NOCS and all the baselines in both the Car and the Plane category, as seen from the average precision curves.

\parahead{Qualitative Results} 
The real-world results in Fig. \ref{fig:real_results} exhibit DRACO's ability of sim2real transfer when solely trained synthetic data, DRACO20K. In Fig. \ref{fig:syn_results} we demonstrate the utility of canonicalization by taking multiple cars having different orientations and canonicalizing them in the NOCS container without performing any form of alignment. We can leverage this form of canonicalization to obtain multiview aggregation of the object effortlessly which is critical to robotics and vision tasks.

\section{Conclusion}
We presented \coolname, a method for dense reconstruction and canonicalization of objects seen in one or more RGB images.
Importantly, \coolname requires only weak supervision from object segmentation mask and camera pose during training, and \emph{only the input RGB images during inference}.
\coolname achieves performance competitive and/or superior compared to fully supervised techniques as well as other semi-supervised methods that use strictly more information.
Future efforts will address multi-object scenes and will further decrease the extent of supervision required for cross-category canonicalization.

\vfill



\bibliographystyle{IEEEtran}
\bibliography{references}

\begin{thebibliography}{10}
\providecommand{\url}[1]{#1}
\csname url@rmstyle\endcsname
\providecommand{\newblock}{\relax}
\providecommand{\bibinfo}[2]{#2}
\providecommand\BIBentrySTDinterwordspacing{\spaceskip=0pt\relax}
\providecommand\BIBentryALTinterwordstretchfactor{4}
\providecommand\BIBentryALTinterwordspacing{\spaceskip=\fontdimen2\font plus
\BIBentryALTinterwordstretchfactor\fontdimen3\font minus
  \fontdimen4\font\relax}
\providecommand\BIBforeignlanguage[2]{{%
\expandafter\ifx\csname l@#1\endcsname\relax
\typeout{** WARNING: IEEEtran.bst: No hyphenation pattern has been}%
\typeout{** loaded for the language `#1'. Using the pattern for}%
\typeout{** the default language instead.}%
\else
\language=\csname l@#1\endcsname
\fi
#2}}

\bibitem{manipulationbook}
R.~M. Murray, Z.~Li, S.~S. Sastry, and S.~S. Sastry, \emph{A mathematical
  introduction to robotic manipulation}.\hskip 1em plus 0.5em minus 0.4em\relax
  CRC press, 1994.

\bibitem{kemp2007challenges}
C.~C. {Kemp}, A.~{Edsinger}, and E.~{Torres-Jara}, ``Challenges for robot
  manipulation in human environments [grand challenges of robotics],''
  \emph{IEEE Robotics Automation Magazine}, vol.~14, no.~1, 2007.

\bibitem{edsinger2007robot}
A.~Edsinger, ``Robot manipulation in human environments,'' 2007.

\bibitem{campbell2010autonomous}
M.~Campbell, M.~Egerstedt, J.~P. How, and R.~M. Murray, ``Autonomous driving in
  urban environments: approaches, lessons and challenges,'' \emph{Philosophical
  Transactions of the Royal Society A: Mathematical, Physical and Engineering
  Sciences}, vol. 368, no. 1928, 2010.

\bibitem{thrun2006stanley}
S.~Thrun, M.~Montemerlo, H.~Dahlkamp, D.~Stavens, A.~Aron, J.~Diebel, P.~Fong,
  J.~Gale, M.~Halpenny, G.~Hoffmann, \emph{et~al.}, ``Stanley: The robot that
  won the darpa grand challenge,'' \emph{Journal of field Robotics}, vol.~23,
  no.~9, 2006.

\bibitem{dsdnet}
W.~Zeng, S.~Wang, R.~Liao, Y.~Chen, B.~Yang, and R.~Urtasun, ``Dsdnet: Deep
  structured self-driving network,'' \emph{Proceedings of the European
  Conference on Computer Vision}, 2020.

\bibitem{goodrich2008human}
M.~A. Goodrich and A.~C. Schultz, \emph{Human-robot interaction: a
  survey}.\hskip 1em plus 0.5em minus 0.4em\relax Now Publishers Inc, 2008.

\bibitem{manuelli2019kpam}
L.~Manuelli, W.~Gao, P.~Florence, and R.~Tedrake, ``kpam: Keypoint affordances
  for category-level robotic manipulation,'' \emph{arXiv:1903.06684}, 2019.

\bibitem{wang20206_6pack}
C.~Wang, R.~Mart{\'\i}n-Mart{\'\i}n, D.~Xu, J.~Lv, C.~Lu, L.~Fei-Fei,
  S.~Savarese, and Y.~Zhu, ``6-pack: Category-level 6d pose tracker with
  anchor-based keypoints,'' in \emph{IEEE International Conference on Robotics
  and Automation (ICRA)}, 2020.

\bibitem{burchfiel2017bayesian}
B.~Burchfiel and G.~D. Konidaris, ``Bayesian eigenobjects: A unified framework
  for 3d robot perception.'' in \emph{Robotics Science and Systems}, 2017.

\bibitem{nocs}
H.~Wang, S.~Sridhar, J.~Huang, J.~Valentin, S.~Song, and L.~J. Guibas,
  ``Normalized object coordinate space for category-level 6d object pose and
  size estimation,'' in \emph{Proceedings of Computer Vision and Pattern
  Recognition}, 2019.

\bibitem{c3dpo}
D.~Novotny, N.~Ravi, B.~Graham, N.~Neverova, and A.~Vedaldi, ``C3dpo: Canonical
  3d pose networks for non-rigid structure from motion,'' in \emph{Proceedings
  of International Conference on Computer Vision}, 2019.

\bibitem{articulatedpose}
X.~Li, H.~Wang, L.~Yi, L.~J. Guibas, A.~L. Abbott, and S.~Song,
  ``Category-level articulated object pose estimation,'' in \emph{Proceedings
  of Computer Vision and Pattern Recognition}, 2020.

\bibitem{xnocs}
S.~Sridhar, D.~Rempe, J.~Valentin, B.~Sofien, and L.~J. Guibas, ``Multiview
  aggregation for learning category-specific shape reconstruction,'' in
  \emph{Neural Information Processing Systems}, 2019.

\bibitem{kar2015cvpr}
A.~Kar, S.~Tulsiani, J.~Carreira, and J.~Malik, ``Category-specific object
  reconstruction from a single image,'' in \emph{Proceedings of Computer Vision
  and Pattern Recognition}, 2015.

\bibitem{pandya2015icra}
H.~{Pandya}, K.~{Madhava Krishna}, and C.~V. {Jawahar}, ``Servoing across
  object instances: Visual servoing for object category,'' in \emph{IEEE
  International Conference on Robotics and Automation (ICRA)}, 2015.

\bibitem{km2017icra}
J.~Krishna~Murthy, G.~Sai~Krishna, F.~Chhaya, and K.~Madhava~Krishna,
  ``Reconstructing vehicles from a single image: Shape priors for road scene
  understanding,'' \emph{IEEE International Conference on Robotics and
  Automation (ICRA)}, 2017.

\bibitem{godard2017unsupervised}
C.~Godard, O.~Mac~Aodha, and G.~J. Brostow, ``Unsupervised monocular depth
  estimation with left-right consistency,'' in \emph{Proceedings of Computer
  Vision and Pattern Recognition}, 2017.

\bibitem{densevo}
C.~Kerl, J.~Sturm, and D.~Cremers, ``Dense visual slam for rgb-d cameras,'' in
  \emph{IEEE International Conference on Intelligent Robots and Systems
  (IROS)}, 2013.

\bibitem{kinectfusion}
R.~A. {Newcombe}, S.~{Izadi}, O.~{Hilliges}, D.~{Molyneaux}, D.~{Kim}, A.~J.
  {Davison}, P.~{Kohi}, J.~{Shotton}, S.~{Hodges}, and A.~{Fitzgibbon},
  ``Kinectfusion: Real-time dense surface mapping and tracking,'' in \emph{2011
  10th IEEE International Symposium on Mixed and Augmented Reality}, 2011.

\bibitem{florence2018dense}
P.~R. Florence, L.~Manuelli, and R.~Tedrake, ``Dense object nets: Learning
  dense visual object descriptors by and for robotic manipulation,''
  \emph{International Conference on Robot Learning}, 2018.

\bibitem{newcombe2015dynamicfusion}
R.~A. Newcombe, D.~Fox, and S.~M. Seitz, ``Dynamicfusion: Reconstruction and
  tracking of non-rigid scenes in real-time,'' in \emph{Proceedings of Computer
  Vision and Pattern Recognition}, 2015.

\bibitem{garg2016unsupervised}
R.~Garg, V.~K. Bg, G.~Carneiro, and I.~Reid, ``Unsupervised cnn for single view
  depth estimation: Geometry to the rescue,'' in \emph{Proceedings of the
  European Conference on Computer Vision}, 2016.

\bibitem{zhou2017unsupervised}
T.~Zhou, M.~Brown, N.~Snavely, and D.~G. Lowe, ``Unsupervised learning of depth
  and ego-motion from video,'' in \emph{Proceedings of Computer Vision and
  Pattern Recognition}, 2017.

\bibitem{monodepth2}
C.~Godard, O.~{Mac Aodha}, M.~Firman, and G.~J. Brostow, ``Digging into
  self-supervised monocular depth prediction,'' in \emph{Proceedings of
  International Conference on Computer Vision}, 2019.

\bibitem{packnet}
V.~Guizilini, R.~Ambrus, S.~Pillai, A.~Raventos, and A.~Gaidon, ``3d packing
  for self-supervised monocular depth estimation,'' in \emph{Proceedings of
  Computer Vision and Pattern Recognition}, 2020.

\bibitem{chen2020learning}
D.~Chen, J.~Li, Z.~Wang, and K.~Xu, ``Learning canonical shape space for
  category-level 6d object pose and size estimation,'' in \emph{Proceedings of
  Computer Vision and Pattern Recognition}, 2020.

\bibitem{manhardt2020cps}
F.~Manhardt, M.~Nickel, S.~Meier, L.~Minciullo, and N.~Navab, ``Cps:
  Class-level 6d pose and shape estimation from monocular images,'' in
  \emph{arXiv}, 2020.

\bibitem{parkhiya2018icra}
P.~Parkhiya, R.~Khawad, J.~Krishna~Murthy, K.~Madhava~Krishna, and B.~Bhowmick,
  ``Constructing category-specific models for monocular object slam,''
  \emph{IEEE International Conference on Robotics and Automation (ICRA)}, 2018.

\bibitem{mu2016iros}
B.~Mu, S.-Y. Liu, L.~Paull, J.~Leonard, and J.~P. How, ``Slam with objects
  using a nonparametric pose graph,'' \emph{IEEE International Conference on
  Intelligent Robots and Systems (IROS)}, 2016.

\bibitem{nicholson2018quadricslam}
L.~Nicholson, M.~Milford, and N.~S{\"u}nderhauf, ``Quadricslam: Dual quadrics
  from object detections as landmarks in object-oriented slam,'' \emph{IEEE
  Robotics and Automation Letters}, 2018.

\bibitem{mccormac2018fusion++}
J.~McCormac, R.~Clark, M.~Bloesch, A.~Davison, and S.~Leutenegger, ``Fusion++:
  Volumetric object-level slam,'' in \emph{International Conference on 3D
  Vision}, 2018.

\bibitem{li2020category}
X.~Li, H.~Wang, L.~Yi, L.~J. Guibas, A.~L. Abbott, and S.~Song,
  ``Category-level articulated object pose estimation,'' in \emph{Proceedings
  of Computer Vision and Pattern Recognition}, 2020.

\bibitem{kulkarni2020articulation}
N.~Kulkarni, A.~Gupta, D.~F. Fouhey, and S.~Tulsiani, ``Articulation-aware
  canonical surface mapping,'' in \emph{Proceedings of Computer Vision and
  Pattern Recognition}, 2020.

\bibitem{zakharov2020autolabeling}
S.~Zakharov, W.~Kehl, A.~Bhargava, and A.~Gaidon, ``Autolabeling 3d objects
  with differentiable rendering of sdf shape priors,'' in \emph{Proceedings of
  Computer Vision and Pattern Recognition}, 2020.

\bibitem{beyondpixels}
S.~Sharma, J.~A. Ansari, J.~Krishna~Murthy, and K.~M. Krishna, ``Beyond pixels:
  Leveraging geometry and shape cues for multi-object tracking,'' \emph{IEEE
  International Conference on Robotics and Automation (ICRA)}, 2018.

\bibitem{kulkarni2019canonical}
N.~Kulkarni, A.~Gupta, and S.~Tulsiani, ``Canonical surface mapping via
  geometric cycle consistency,'' in \emph{Proceedings of the IEEE International
  Conference on Computer Vision}, 2019.

\bibitem{lei2020pix2surf}
J.~Lei, S.~Sridhar, P.~Guerrero, M.~Sung, N.~Mitra, and L.~J. Guibas,
  ``Pix2surf: Learning parametric 3d surface models of objects from images,''
  in \emph{Proceedings of the European Conference on Computer Vision}, 2020.

\bibitem{resnet}
K.~{He}, X.~{Zhang}, S.~{Ren}, and J.~{Sun}, ``Deep residual learning for image
  recognition,'' in \emph{Proceedings of Computer Vision and Pattern
  Recognition}, 2016.

\bibitem{sfmlearner}
T.~Zhou, M.~Brown, N.~Snavely, and D.~G. Lowe, ``Unsupervised learning of depth
  and ego-motion from video,'' in \emph{Proceedings of Computer Vision and
  Pattern Recognition}, 2017.

\bibitem{elu}
D.-A. {Clevert}, T.~{Unterthiner}, and S.~{Hochreiter}, ``Fast and accurate
  deep network learning by exponential linear units (elus),'' in \emph{arXiv},
  2015.

\bibitem{instancenorm}
D.~Ulyanov, A.~Vedaldi, and V.~S. Lempitsky, ``Instance normalization: The
  missing ingredient for fast stylization,'' \emph{arXiv}, 2016.

\bibitem{ssim}
H.~{Zhao}, O.~{Gallo}, I.~{Frosio}, and J.~{Kautz}, ``Loss functions for image
  restoration with neural networks,'' \emph{IEEE Transactions on Computational
  Imaging}, vol.~3, no.~1, 2017.

\bibitem{Johnson2016Perceptual}
J.~Johnson, A.~Alahi, and L.~Fei-Fei, ``Perceptual losses for real-time style
  transfer and super-resolution,'' in \emph{Proceedings of the European
  Conference on Computer Vision}, 2016.

\bibitem{vgg}
K.~Simonyan and A.~Zisserman, ``{Very deep convolutional networks for
  large-scale image recognition},'' \emph{arXiv preprint arXiv:1409.1556},
  2014.

\bibitem{adam}
D.~P. Kingma and J.~Ba, ``Adam: {A} method for stochastic optimization,'' in
  \emph{International Conference on Learning Representations}, 2015.

\bibitem{chang2015shapenet}
A.~X. Chang, T.~Funkhouser, L.~Guibas, P.~Hanrahan, Q.~Huang, Z.~Li,
  S.~Savarese, M.~Savva, S.~Song, H.~Su, \emph{et~al.}, ``Shapenet: An
  information-rich 3d model repository,'' \emph{arXiv:1512.03012}, 2015.

\bibitem{lin2015microsoft}
T.-Y. Lin, M.~Maire, S.~Belongie, L.~Bourdev, R.~Girshick, J.~Hays, P.~Perona,
  D.~Ramanan, C.~L. Zitnick, and P.~Dollár, ``Microsoft coco: Common objects
  in context,'' 2015.

\bibitem{you2020keypointnet}
Y.~You, Y.~Lou, C.~Li, Z.~Cheng, L.~Li, L.~Ma, C.~Lu, and W.~Wang,
  ``Keypointnet: A large-scale 3d keypoint dataset aggregated from numerous
  human annotations,'' \emph{arXiv:2002.12687}, 2020.

\bibitem{umeyama1991least}
S.~Umeyama, ``Least-squares estimation of transformation parameters between two
  point patterns,'' \emph{IEEE Transactions on Pattern Analysis \& Machine
  Intelligence}, no.~4, 1991.

\bibitem{shape_priors_kp_net}
J.~K. {Murthy}, S.~{Sharma}, and K.~M. {Krishna}, ``Shape priors for real-time
  monocular object localization in dynamic environments,'' in \emph{IEEE
  International Conference on Intelligent Robots and Systems (IROS)}, 2017.

\end{thebibliography}

\end{document}